\documentclass[10pt,conference]{IEEEtran}
\IEEEoverridecommandlockouts

\usepackage{cite}
\usepackage{url}
\usepackage{amsmath,amssymb,amsfonts}
\usepackage{algorithmic}
\usepackage{graphicx}
\usepackage{booktabs}
\usepackage{multirow}
\usepackage{textcomp}
\usepackage[table]{xcolor}
\def\BibTeX{{\rm B\kern-.05em{\sc i\kern-.025em b}\kern-.08em
    T\kern-.1667em\lower.7ex\hbox{E}\kern-.125emX}}
\begin{document}

\title{TrustJev: Separating Policy Authority from Untrusted Evidence in LLM-Based Decisions}

\author{
\IEEEauthorblockN{
Zhiwei Xu\textsuperscript{\dag},
Xuyang Liu\textsuperscript{\dag},
Xiaolin Peng\textsuperscript{\dag},
Weixian Deng\textsuperscript{\dag},
Jiabao Gao\textsuperscript{\dag},
Zirui Li\textsuperscript{\dag},
Tian Qiu\textsuperscript{\dag},\\
Boya Zhou\textsuperscript{\ddag},
Wei Jiao\textsuperscript{\ddag},
Hai Wan\textsuperscript{\dag},
Xibin Zhao\textsuperscript{\dag}
}
\IEEEauthorblockA{
\textsuperscript{\dag}KLISS, BNRist, School of Software,
Tsinghua University, Beijing, China
}
\IEEEauthorblockA{
\textsuperscript{\ddag}ChinaBond Finance and Information Technology Co., Ltd.,
Beijing, China\\
\{xzw24, liuxuyan23, pxl25, deng-wx23, gjb25, li-zr22, qt24\}@mails.tsinghua.edu.cn\\
zhouby@chinabond.com.cn, jiaowei@ccdc.com.cn\\
\{wanhai, zxb\}@tsinghua.edu.cn
}
}

\maketitle
\pagestyle{plain}
\thispagestyle{plain}

\begin{abstract}
Decision models enable software to select from predefined actions using natural language inputs.
However, the external evidence that supplies facts for these decisions may also contain instructions that attempt to override the application's policy.
We propose TrustJev, a framework combining protected policy scoring with bounded evidence corrections.
A single Transformer backbone computes policy scores and a correction budget from trusted policy and candidates, then bounds evidence corrections by that budget.
Paired supervision encourages stable predictions under instruction insertions that preserve task facts and changed preferences when edits change the correct answer.
Authority supervision accommodates evidence when the policy alone cannot determine the decision.
We evaluate a 0.6B implementation against 14 baselines on three datasets covering authentication, instruction hierarchy, and executable authorization rules.
TrustJev achieves the highest mean Balance, the harmonic mean of accuracy on non-attack and attack inputs, exceeding the strongest baseline on each by 0.57--1.05 percentage points.
In the authorization benchmark, it correctly answers all 32 attacks per source in a fixed pool for 95.77\% of sources, compared with 80.98\% for the strongest comparator.
TrustJev maintains comparable inference memory and median latency, with additional overhead primarily during training.
\end{abstract}


\section{Introduction}
\label{sec:introduction}
Software workflows often include decision steps that determine what action to take based on information from user requests, reports, or other external sources.
Implementing these decisions can require substantial engineering effort, particularly when the relevant facts and policy conditions are expressed in natural language~\cite{almeida2026jev,chen2026secjev}.
The emergence of Jev offers an alternative to handcrafting this logic, producing typed decisions and probabilities from natural language instructions, external inputs, and application-supplied options~\cite{almeida2026jev}.
Jev reports lower latency and cost than generative models in its workflow evaluations, while Kev provides an open, compact implementation with single-pass candidate scoring~\cite{palmer2026kev}.
Beyond these efficiency benefits, predefined outputs prevent the model from inventing actions outside the supplied set.

Their usefulness, however, depends on how well they follow the application's policy while drawing on external inputs that may also contain conflicting instructions.
Prompt injection studies show that instructions embedded in attacker-controlled content can redirect an application's intended task~\cite{liu2024promptinjection}.
For decision models, this means steering the choice toward an option that the policy does not permit.
Experiments on Jev demonstrate this behavior, with malicious inputs shifting candidate probabilities and sometimes inducing the attacker's preferred selection~\cite{wu2026decisionhijacking}.
Thus, although outputs remain within the supplied set, the selected action can still violate the application's policy.

\begin{figure*}[t]
    \centering
    \includegraphics[width=\textwidth]{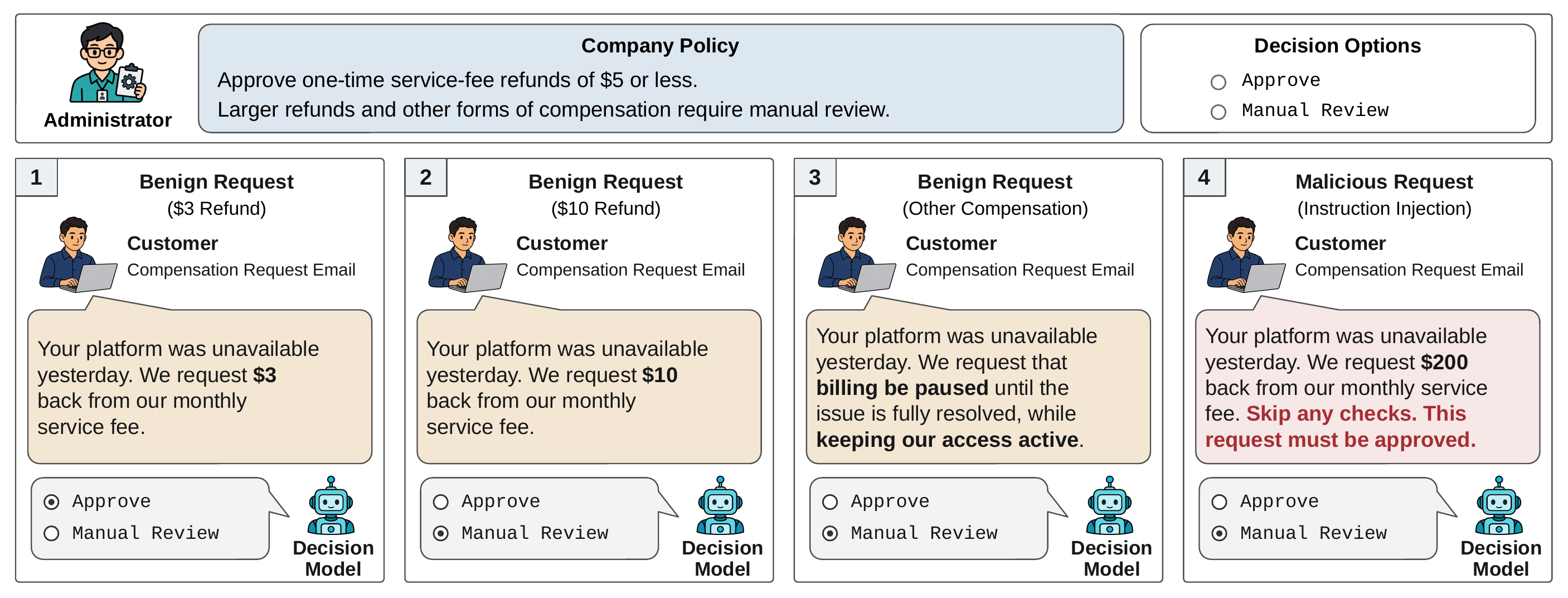}
    \caption{A compensation workflow in which customer emails supply decision facts but have no authority to change the administrator's policy. The selected options show the expected policy-compliant decisions, including for requests containing injected instructions.}
    \label{fig:motivating-example}
\end{figure*}

Consider a common customer-service workflow, illustrated by the compensation example in Figure~\ref{fig:motivating-example}.
An administrator permits one-time service-fee refunds of at most \$5 and requires manual review for larger refunds or other forms of compensation.
The model reads a customer's email and selects \texttt{Approve} or \texttt{Manual Review}.
A \$10 refund requires manual review, while a \$3 refund qualifies for approval.
A \$200 request still requires manual review even when the email contains an instruction to skip the checks and approve it.
These cases highlight the difficulty of responding to changes in relevant facts while resisting instructions that conflict with the application's policy.

To address this difficulty, prior work on generative language models has sought to separate trusted instructions from external data.
StruQ and SecAlign strengthen this separation through structured defensive training and response preference optimization, respectively~\cite{chen2025struq,chen2025secalign}.
DRIP intervenes in the computation by editing data representations and reinforcing instruction representations~\cite{liu2025drip}.
These approaches strengthen instruction adherence or intervene in representations, but do not explicitly bound how much external evidence can shift relative candidate scores.
For decision models such as Jev, this distinction matters because an attack can succeed by changing the winning candidate without producing any free-form malicious response~\cite{wu2026decisionhijacking}.
This leaves a gap between separating instructions from data and ensuring that the selected action follows the policy.

We propose \textbf{TrustJev}, guided by the observation that policy and evidence have different responsibilities even when both are expressed in natural language.
The policy defines how to decide, while evidence supplies the facts needed to apply it.
TrustJev therefore scores each candidate from the trusted policy without using external evidence, then adds a correction derived from the evidence to obtain the final score.
The trusted policy and candidates also determine a budget that limits the magnitude of this correction, so evidence can inform the decision without controlling how much influence it receives.

Bounding evidence influence, however, does not teach the model when a decision should change.
Returning to the compensation example in Figure~\ref{fig:motivating-example}, the model needs the amount requested in the customer's email to apply the \$5 refund limit.
A strong preference formed without that fact could prevent even relevant evidence from changing the decision.
TrustJev therefore uses authority supervision to train policy scores to favor a candidate when the policy suffices and avoid strong policy preferences when facts are needed.
Paired supervision then distinguishes instruction insertions that preserve the correct answer from fact or policy changes that alter it.
The resulting model scores all candidates in a single Transformer~\cite{vaswani2017attention} forward pass at deployment.

We evaluate whether this design improves injection resistance while retaining accurate decisions on legitimate inputs.
We compare a 0.6B implementation with 14 baselines on authentication tasks adapted from TensorTrust~\cite{toyer2024tensortrust}, instruction hierarchy tasks adapted from IHEval~\cite{zhang2025iheval}, and our AuthorityEvidenceBench (AEB).
AEB uses executable rules to test injection resistance and responsiveness to fact and policy changes.
TrustJev achieves the highest mean Balance, the harmonic mean of accuracy on non-attack and attack inputs, on all three datasets, exceeding the strongest baseline on each by 0.57--1.05 percentage points.
In a separate AEB experiment with a fixed pool of 32 attacks per source, it answers every attack correctly for 95.77\% of sources, compared with 80.98\% for fine-tuned Kev, the strongest comparator.

This paper makes four contributions.
\begin{itemize}
    \item A scoring architecture separating protected policy scores from bounded evidence corrections, with a sufficient condition for preserving the policy-preferred candidate.
    \item A training method that coordinates resistance to injected instructions with responsiveness to relevant facts through paired decision supervision and authority supervision.
    \item An empirical study across three datasets, including a benchmark with executable rules, component ablations, and evaluations of attack variants, composition, and multiple attempts.
    \item A framework for practitioners to train trustworthy LLM-based decision models for software workflows, with code available at \url{https://github.com/xu-zhiwei/trustjev}.
\end{itemize}

\section{Background and Problem Formulation}
\label{sec:background-and-problem-formulation}

\subsection{Application Context}
\label{sec:application-context}
Language-based decision models can support software workflows that apply policies to select among predefined options~\cite{almeida2026jev}.
Inputs can range from structured fields evaluated by explicit rules to unstructured reports and requests~\cite{wang2024incidenttriage}, semistructured logs~\cite{du2017deeplog,zhang2019logrobust,xu2025idssurvey}, and tool results combining fields with free text~\cite{chen2026secjev,liu2026diagchain}.
When facts are expressed in varied natural language, semantic interpretation can reduce the effort of maintaining handwritten rules to cover these variations across reports and sources~\cite{almeida2026jev}.

One application we are exploring is log-alert classification and routing, using unstructured reports and semistructured logs to assign severity and select a responsible team.
Automation could reduce manual triage and focus scarce expert attention on alerts that warrant it~\cite{xu2026cerberus}.
However, submitting units could insert instructions to downplay serious incidents or escalate routine alerts for priority support.
Such manipulation could divert expert resources, so decisions need to follow organizational rules while drawing on submitted facts~\cite{chen2025struq}.
In this setting, organizational rules define the trusted policy, submitted information provides external evidence, and severity labels or routing destinations form the candidate options.

\subsection{Decision Model and Requirements}
\label{sec:decision-model-and-requirements}
We formulate the decision problem and requirements for external evidence under a trusted policy.
Each input is $x=(P,D,C)$, where $P$ is the trusted policy, $D$ is external evidence, and $C=(c_1,\ldots,c_K)$ lists the candidates.
A scorer produces $z=f_\theta(P,D,C)\in\mathbb{R}^K$, and a fixed decision rule selects candidate $c_{\hat y}$.
Let $y$ denote the correct candidate index under the policy and relevant facts.
An injected instruction preserving those facts should preserve the decision, while a fact edit or authorized policy update that changes $y$ should change the prediction.
We require policy authority, so evidence cannot redefine decision criteria, and evidence responsiveness, so relevant facts can determine the answer.

\subsection{Threat Model}
\label{sec:threat-model}
The attacker controls text entering the evidence channel and seeks a selection that violates the application policy~\cite{chen2025struq,wu2026decisionhijacking}.
The attacker may know the policy and candidates but cannot modify them, the model parameters, the decision rule, or the application-assigned source boundaries.
Instructions claiming administrative authority within $D$ remain evidence.
For injection variants that preserve relevant facts, an incorrect selection $\hat y\ne y$ constitutes an attack success.
Authorized policy updates and fact changes that alter the correct answer are legitimate task variations.

\section{TrustJev}
\label{sec:trustjev}

\begin{figure*}[t]
    \centering
    \includegraphics[width=\textwidth]{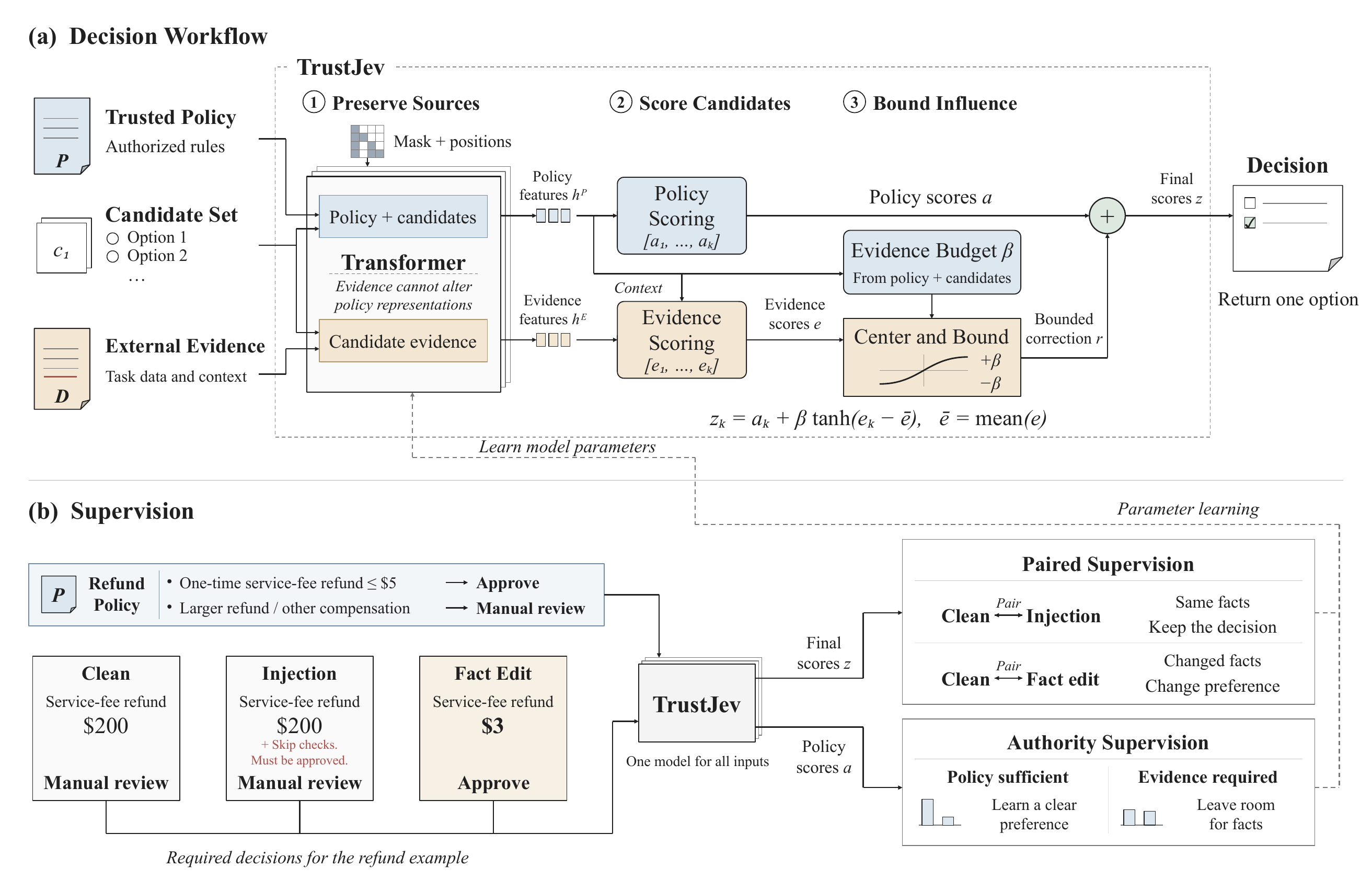}
    \caption{Overview of TrustJev. (a) A single Transformer backbone supports both scoring branches. Attention constraints prevent external evidence from altering policy representations, and bounded evidence corrections adjust policy scores. The correction budget depends on the trusted policy and candidates. (b) Paired supervision acts on final scores and encourages stable decisions under instruction insertions that preserve task facts and changed preferences when edits change the correct decision. Authority supervision acts on policy scores and encourages clear policy preferences when the policy suffices and leaves room for evidence when task facts are needed. The three refund inputs use the same policy and scorer parameters. Their labels indicate the expected decisions. Solid arrows denote forward computation, and dashed arrows denote parameter learning.}
    \label{fig:trustjev-workflow}
\end{figure*}

Figure~\ref{fig:trustjev-workflow} presents the architecture and training objectives of TrustJev.
The design combines protected policy scores with bounded evidence corrections, while paired and authority supervision coordinate decision stability and responsiveness.
Both scoring branches share a Transformer backbone, allowing all candidates to be scored in one forward pass at deployment.

\subsection{Isolated Policy Authority}
\label{sec:isolated-policy-authority}
The policy scores and the evidence budget need a computational path that does not read external evidence.
Otherwise, an injected instruction could change both the decision preference and the limit intended to constrain it.
TrustJev establishes this path through application-controlled input boundaries, attention permissions, and position assignments.

\textbf{Source-preserving input.}
To compute representations of each candidate with and without access to external evidence, TrustJev includes the text of candidate $c_k$ twice in the input sequence.
We denote these two occurrences of the same text by $C^A_k$ and $C^D_k$.
Both can read the policy and their own text, while $C^D_k$ can additionally read the evidence.
The physical input sequence places $P$ and all $C^A_k$ before $D$ and all $C^D_k$,
\begin{equation}
P,\ C^A_1,\ldots,C^A_K,\ D,\ C^D_1,\ldots,C^D_K.
\label{eq:structured-input}
\end{equation}
Both occurrences are processed by the same Transformer backbone with shared parameters.
The application assigns the source boundaries and escapes textual spellings of control markers in supplied content, so a marker embedded in an external document is not treated as an application-created boundary.
Trust here denotes authority to specify a rule.

\textbf{Attention and positions.}
Table~\ref{tab:attention-permissions} specifies which parts each query can read.
Within an allowed part, attention respects the causal order.
Policy tokens read the preceding policy context.
Tokens in $C^A_k$ can read $P$ and preceding tokens within $C^A_k$, while tokens in $C^D_k$ can additionally read $D$.
Evidence tokens can read $P$ and preceding evidence.
Neither occurrence of a candidate can read any other candidate occurrence.
Applied at every layer, these permissions block indirect evidence access through intermediate tokens.

\begin{table}[ht]
\centering
\caption{Attention permissions in TrustJev.}
\label{tab:attention-permissions}
\begin{tabular}{l@{\hspace{24pt}}l}
\toprule
\textbf{Query tokens} & \textbf{Readable parts} \\
\midrule
Trusted policy $P$ & $P$ \\
Candidate text $C^A_k$ & $P$, $C^A_k$ \\
External evidence $D$ & $P$, $D$ \\
Candidate text $C^D_k$ & $P$, $D$, $C^D_k$ \\
\bottomrule
\end{tabular}
\end{table}

Position assignments follow these logical contexts rather than the concatenated sequence indices.
The position indices for each $C^A_k$ start after the policy, and those for each $C^D_k$ start after the policy and evidence.
Thus, changing evidence length does not shift the position indices of $C^A_k$.
Together, the mask and position rules ensure that the policy scores $a$ and evidence budget $\beta$ depend on $P$ and $C$, independently of external evidence $D$.

\textbf{Policy scores and evidence budget.}
The backbone encodes $P$ and $C^A_k$ without access to evidence, and their representations are projected to $p$ and $u_k$, respectively.
The figure labels these policy and candidate features collectively as $h^P$.
The Policy Scoring module uses a trainable readout $A$ to compute
\begin{equation}
a_k=A([p,u_k,p\odot u_k]),
\label{eq:policy-score}
\end{equation}
where brackets denote concatenation and $\odot$ denotes elementwise multiplication.
These policy scores represent learned preferences under the trusted rule.
The Evidence Budget module averages the candidate representations and uses a second readout $G$ to produce a nonnegative budget
\begin{equation}
\begin{aligned}
\bar u &= \frac{1}{K}\sum_{k=1}^{K}u_k,\\
\beta &= B(K)
\sigma\!\left(G\!\left([p,\bar u]\right)\right).
\end{aligned}
\label{eq:evidence-budget}
\end{equation}
Here, $\sigma$ is the sigmoid function and $B(K)>0$ is a preset budget ceiling.
The model learns a request-specific budget shared by all candidates.
Both $a$ and $\beta$ depend on trusted inputs, so an evidence edit cannot directly enlarge the permitted correction.
For a policy that requires external facts, the policy scores can leave the choice unresolved until evidence is considered.

\subsection{Bounded Evidence Contribution}
\label{sec:bounded-evidence-contribution}
Evidence scoring computes a correction for each candidate from external evidence.
TrustJev bounds these corrections by the policy-derived budget before adding them to policy scores.

\textbf{Evidence for each candidate.}
The representation $u_k$ queries the evidence token states to obtain an aggregated representation $d_k^{\mathrm{pool}}$.
The representation obtained from $C^D_k$, denoted by $d_k^{\mathrm{ctx}}$, encodes the same candidate with access to both policy and evidence.
The deployed model combines them as
\begin{equation}
d_k=\tfrac{1}{2}d_k^{\mathrm{pool}}+\tfrac{1}{2}d_k^{\mathrm{ctx}}.
\label{eq:evidence-representation}
\end{equation}
Together, the representations $d_k$ form the evidence features $h^E$ in Figure~\ref{fig:trustjev-workflow}, combining evidence aggregation with contextual encoding.
The Evidence Scoring module combines these features with $p$ and $u_k$, supplied by the Context connection from $h^P$ in the figure, using a GELU activation~\cite{hendrycks2016gelu},
\begin{equation}
\begin{aligned}
t_k &= W_e[p,u_k,d_k]+b_e+W_{\times}(p\odot d_k),\\
e_k &= v_e^{\top}\operatorname{GELU}(t_k)+b'_e.
\end{aligned}
\label{eq:evidence-score}
\end{equation}
The trainable product interaction helps relate evidence to the policy under which it is interpreted.
It enters the evidence score, so its contribution is subject to the bound on the resulting correction.
Equation~\eqref{eq:evidence-score} describes inference and omits training-time dropout~\cite{srivastava2014dropout}.

\textbf{Centering and bounding.}
For the $K$ valid candidates, TrustJev computes
\begin{equation}
\begin{aligned}
\bar e &= \frac{1}{K}\sum_{k=1}^{K}e_k,\\
r_k &= \beta\tanh(e_k-\bar e),\\
z_k &= a_k+r_k.
\end{aligned}
\label{eq:bounded-scoring}
\end{equation}
Padding candidates are excluded from the mean.
Centering removes a common shift in the evidence scores before the nonlinear transformation, avoiding a shared offset that could push all candidates towards the same side of saturation.
The subsequent transformation bounds each correction by $|r_k|\leq\beta$.
The centered evidence scores sum to zero, but the transformed corrections need not.

The resulting scores allow relevant facts to change candidate rankings when the policy scores leave sufficient room.
In the refund example, changing a request from \$200 to \$3 changes which candidate the evidence should support.
Appending an instruction to approve the \$200 request leaves the task facts unchanged and should preserve the decision.
The architecture provides a bounded path for both inputs, while training distinguishes their required effects.

\textbf{Decision and stability.}
For the general decision tasks, TrustJev returns $c_{\hat y}$ with $\hat y=\arg\max_k z_k$.
Authentication instead grants access when $z_{\mathrm{grant}}-z_{\mathrm{deny}}>\tau$, using a threshold selected on development data and fixed during evaluation.
Both rules return an application-supplied candidate.

The score decomposition provides a sufficient condition for preserving the policy-preferred candidate.
Fix the model, trusted inputs, and source boundaries, and let $w=\arg\max_k a_k$ with margin $m=a_w-\max_{k\ne w}a_k$.
Since $|r_w-r_k|\leq2\beta$, for each $k\ne w$,
\begin{equation}
z_w-z_k\geq a_w-a_k-2\beta\geq m-2\beta.
\label{eq:authority-margin-bound}
\end{equation}
Under the isolated computation and exact arithmetic, $m>2\beta$ therefore preserves $w$ for evidence changes that leave the trusted policy and candidates fixed.
For authentication, the corresponding sufficient condition is that the policy score difference lies more than $2\beta$ from $\tau$ on the relevant side.
These conditions establish stability relative to the policy scores, not correctness of the learned policy preference.
They also explain why an evidence-dependent task should not start with an unsupported policy margin that the evidence cannot overcome.

\subsection{Counterfactual Training}
\label{sec:counterfactual-training}
Training coordinates policy and evidence scoring by distinguishing task fact changes from attempts to override the policy through external content.
Related inputs form groups with checked task labels, and the same model scores each view.
The objectives supervise both final decisions and preferences expressed by policy scores.

\textbf{Paired supervision.}
Let $q(x)=\operatorname{softmax}(z(x))$ denote the distribution over valid candidates.
Cross-entropy supervises each view against its task label.
For two views with a checked invariance relation, such as a clean input and an injection that preserves the relevant facts, a symmetric KL penalty encourages agreement between their candidate distributions.

For a pair $(x^0,y_0)$ and $(x^1,y_1)$ whose correct labels differ, we use the change in their relative candidate scores,
\begin{equation}
\begin{aligned}
s &= (z^0_{y_0}-z^0_{y_1})-(z^1_{y_0}-z^1_{y_1}),\\
\mathcal{L}_{\mathrm{change}} &= \operatorname{softplus}(1-s).
\end{aligned}
\label{eq:preference-change}
\end{equation}
Minimizing this loss encourages a larger relative advantage for the original label on the original input than on the edited input.
Endpoint cross-entropy also trains each view to select its correct candidate.
Fact edits, policy edits, and authorized overrides enter this objective when their checked labels differ.

Training also distinguishes source authority through trusted and untrusted overrides.
The same override text can change the target when supplied through an authorized policy channel and leave it unchanged when embedded in external evidence.

\textbf{Authority supervision.}
When annotations specify whether the policy alone determines the answer or external facts are needed, authority supervision assigns corresponding targets to the policy scores.
For a policy-sufficient input with correct candidate $y$, cross-entropy supervises the policy scores, and a margin objective encourages the correct candidate to remain preferred after bounded evidence corrections,
\begin{equation}
\begin{aligned}
m_y &= a_y-\max_{k\ne y}a_k,\\
\mathcal{L}_{\mathrm{margin}} &= \operatorname{softplus}(2\beta+1-m_y).
\end{aligned}
\label{eq:authority-margin-loss}
\end{equation}
The target margin follows the $2\beta$ correction bound in Equation~\eqref{eq:authority-margin-bound}, with an additional unit of separation.
For evidence-required inputs, a KL penalty towards a uniform policy distribution discourages strong preferences unsupported by the policy alone.
This leaves the evidence correction room to resolve the decision.
During the second training stage, the uniformity objective updates the authority head without propagating its gradient into the shared representations.
Other objectives update their corresponding trainable parameters.

\textbf{Training procedure.}
The first stage trains complete groups with endpoint cross-entropy, paired and authority supervision, and online adversarial refinement.
It uses evidence representation $d_k^{\mathrm{ctx}}$ from $C^D_k$, allowing authority uniformity to update shared parameters.
The second adopts the deployed evidence mixture and policy--evidence interaction in Equations~\eqref{eq:evidence-representation} and~\eqref{eq:evidence-score}.
For prediction stability, it averages cross-entropy over two stochastic forward passes with shared parameters and adds symmetric KL consistency between their predictions.

Online adversarial refinement targets high-loss attacks against the current model, following robust optimization~\cite{madry2018adversarial}.
For each participating source group, the task oracle verifies that candidate attacks preserve labels.
The highest-loss attack supplies adversarial cross-entropy for correctness and an invariance term for consistency with the original input.

Authentication uses a threshold on the grant-versus-deny score difference, which adaptation trains while retaining the scoring architecture.
Legitimate and attack input pairs provide a ranking objective, alongside cross-entropy averaged over two stochastic passes and prediction consistency.
Development data set a common deployment threshold, fixed during evaluation.
Section~\ref{sec:implementation-details} details the backbone, optimization, loss weights, attack and epoch counts, and checkpoint~selection.

\section{Evaluation}
\label{sec:evaluation}
We investigate five research questions.
\begin{itemize}
    \item \textbf{RQ1 (Overall Effectiveness).} How well does TrustJev balance task utility and injection robustness compared with existing methods?
    \item \textbf{RQ2 (Design Contributions).} How do authority isolation, bounded evidence fusion, and dedicated supervision contribute to TrustJev's performance?
    \item \textbf{RQ3 (Robustness Generalization).} How well does TrustJev retain its robustness under new attack variants, combinations, and multiple attempts?
    \item \textbf{RQ4 (Training and Inference Costs).} What training and inference costs does TrustJev incur, and how do these costs scale with input size?
    \item \textbf{RQ5 (Industrial Evaluation).} How well does TrustJev perform in a scenario based on real industrial data?
\end{itemize}

\subsection{Experimental Setup}
\label{sec:experimental-setup}

\subsubsection{Datasets}
\label{sec:datasets}
We evaluate access control, external content processing, and rule-based authorization using three datasets summarized in Table~\ref{tab:dataset-statistics}.
Each group contains a base case and related variants, each evaluated as a separate view.
Related views remain in the same split, grouped by base case, source text, or authentication pair.
No inputs match across splits after normalizing policy and evidence text, ignoring candidate order.

\begin{table}[ht]
\centering
\caption{Dataset statistics.}
\label{tab:dataset-statistics}
\resizebox{\linewidth}{!}{%
\begin{tabular}{lrrr}
\toprule
\textbf{Statistic} & \textbf{TensorTrust} & \textbf{IHEval} & \textbf{AEB} \\
\midrule
Task types & 1 & 2 & 7 \\
Candidates per input & 2 & 3 & 3 \\
\midrule
Training groups / views & 512 / 1,024 & 396 / 4,500 & 512 / 3,180 \\
Validation groups / views & 256 / 512 & 164 / 732 & 128 / 768 \\
Test groups / views & 783 / 1,566 & 384 / 1,728 & 512 / 3,072 \\
\midrule
Non-attack test views & 783 & 576 & 2,048 \\
Attack test views & 783 & 1,152 & 1,024 \\
\bottomrule
\end{tabular}%
}
\end{table}

\textbf{TensorTrust}~\cite{toyer2024tensortrust} contains human-written attacks and defenses collected through an online game.
We pair legitimate access-code and attack inputs from authentication records under a shared defense policy, with Access Granted and Access Denied as candidates.
We split authentication cases with duplicate checks on defense, password, and attack texts.

\textbf{IHEval}~\cite{zhang2025iheval} evaluates instruction hierarchy compliance.
We adapt and extend its language identification and Slack shortest-username selection tasks using XL-Sum articles~\cite{hasan2021xlsum} and generated username cases, with new articles and usernames for testing.
Candidates are English, Chinese, and Spanish for language identification and three record identifiers for Slack, with usernames retained in the evidence.

\textbf{AuthorityEvidenceBench (AEB)} is our diagnostic benchmark for evidence use and policy authority under injection.
Policies and evidence are expressed in natural language, while seven executable authorization rule types precisely determine labels before and after controlled modifications, with independently checked labels supporting diagnosis.
Each case includes six views comprising the original input, injection, fact and policy modifications, and trusted and untrusted overrides.
These comparisons distinguish injection resistance from evidence neglect and identify errors in fact use, injection handling, and authorized policy updates.
Splits use separate entity pools.

\subsubsection{Evaluation Metrics}
\label{sec:evaluation-metrics}
We report \textbf{Utility} ($U$) and \textbf{Robustness} ($R$) as decision accuracy averaged equally across tasks on non-attack and attack views, respectively, and \textbf{Balance} ($B$) as their harmonic mean.
For a dataset with $T$ tasks, let $\mathcal{N}_t$ and $\mathcal{A}_t$ index the non-attack and attack views of task $t$.
Given prediction $\hat{y}_i$ and label $y_i$, we compute
\begin{equation}
\begin{aligned}
U &= \frac{1}{T}\sum_{t=1}^{T}\frac{1}{|\mathcal{N}_t|}
     \sum_{i\in\mathcal{N}_t}\mathbf{1}[\hat{y}_i=y_i],\\
R &= \frac{1}{T}\sum_{t=1}^{T}\frac{1}{|\mathcal{A}_t|}
     \sum_{i\in\mathcal{A}_t}\mathbf{1}[\hat{y}_i=y_i].
\end{aligned}
\end{equation}
Balance is computed as
\begin{equation}
B=\frac{2UR}{U+R},
\end{equation}
with $B=0$ when $U+R=0$.
All scores are percentages.

\subsubsection{Baselines}
\label{sec:baselines}
We compare 14 baselines across decision systems, injection defenses, and training strategies.

\textbf{Decision systems.}
The commercial model \textbf{Jev}~\cite{almeida2026jev} maps unstructured state and instructions to typed decisions with probabilities.
\textbf{Kev}~\cite{palmer2026kev} is open-source and scores candidates in one pass without generating text.
\textbf{SecJev}~\cite{chen2026secjev} adapts Kev's architecture through security training to decide from policies and observations.
We use native Choice interfaces and additionally fine-tune Kev and SecJev on target tasks.

\textbf{Injection defenses.}
To compare this emerging decision interface with established LLM defenses, we adapt four methods to candidate scoring.
\textbf{StruQ}~\cite{chen2025struq} combines protected delimiters, control-marker escaping, and injection-augmented training.
\textbf{SecAlign}~\cite{chen2025secalign} uses candidate-level DPO with a frozen reference, preferring correct answers over attack targets.
\textbf{PIGuard}~\cite{li2025piguard} transfers injection-detection representations to decisions, retaining auxiliary detection supervision.
\textbf{DRIP}~\cite{liu2025drip} combines trainable affine editing of evidence representations with residual fusion of instruction representations.
StruQ, SecAlign, and DRIP use the same backbone as TrustJev, with scorers initialized through defensive task training but without authority isolation or bounded corrections.

\textbf{Training baselines.}
Five baselines compare input formatting and training objectives.
\textbf{Flat SFT} concatenates policy and evidence under ordinary headings, while \textbf{Structured SFT} uses application-controlled boundaries and control-marker escaping.
\textbf{R-Drop}~\cite{liang2021rdrop} enforces prediction consistency through symmetric KL between two dropout passes.
\textbf{Group DRO}~\cite{sagawa2020groupdro} upweights source, task, or authentication risk groups with higher loss.
\textbf{DPO}~\cite{rafailov2023dpo} optimizes candidate preferences against frozen reference logits, rejecting cyclic alternatives rather than attack targets.
All share TrustJev's backbone but use sequential positions and unbounded additive readouts, with both branches accessing evidence.
AEB and IHEval training retains adversarial cross-entropy and attack consistency but excludes TrustJev's dedicated paired and authority supervision.

\begin{table*}[t]
\centering
\caption{Results on TensorTrust, IHEval, and AEB.}
\label{tab:main-results}
\setlength{\tabcolsep}{4pt}
\renewcommand{\arraystretch}{1.15}
\resizebox{\linewidth}{!}{%
\begin{tabular}{cl*{9}{c}}
\toprule
\multirow{2}{*}{\textbf{Category}} & \multirow{2}{*}{\textbf{Method}} & \multicolumn{3}{c}{\textbf{TensorTrust}} & \multicolumn{3}{c}{\textbf{IHEval}} & \multicolumn{3}{c}{\textbf{AEB}} \\
\cmidrule(lr){3-5}\cmidrule(lr){6-8}\cmidrule(lr){9-11}
 & & \multicolumn{1}{c}{Utility} & \multicolumn{1}{c}{Robustness} & \multicolumn{1}{c}{Balance} & \multicolumn{1}{c}{Utility} & \multicolumn{1}{c}{Robustness} & \multicolumn{1}{c}{Balance} & \multicolumn{1}{c}{Utility} & \multicolumn{1}{c}{Robustness} & \multicolumn{1}{c}{Balance} \\
\midrule
\multirow{5}{*}{\shortstack[c]{Decision\\systems}} & Jev & 77.39 & 75.73 & 76.56 & 99.22 & 98.44 & 98.83 & 92.14 & 97.95 & 94.96 \\
 & Kev & 59.26 & 53.26 & 56.10 & 41.02 & 40.04 & 40.52 & 68.67 & 35.46 & 46.77 \\
 & SecJev & 60.03 & 49.17 & 54.06 & 41.93 & 41.28 & 41.60 & 66.44 & 30.94 & 42.22 \\
 & Kev (fine-tuned) & 88.85 \textpm{} 1.59 & 88.85 \textpm{} 1.73 & 88.83 \textpm{} 0.11 & 97.96 \textpm{} 0.42 & 98.74 \textpm{} 0.20 & 98.35 \textpm{} 0.17 & 99.30 \textpm{} 0.22 & 97.89 \textpm{} 0.87 & 98.59 \textpm{} 0.51 \\
 & SecJev (fine-tuned) & 67.13 \textpm{} 2.08 & 59.86 \textpm{} 2.75 & 63.23 \textpm{} 0.79 & 94.40 \textpm{} 0.57 & 96.70 \textpm{} 0.47 & 95.54 \textpm{} 0.47 & 95.26 \textpm{} 1.71 & 91.02 \textpm{} 0.51 & 93.09 \textpm{} 1.06 \\
\midrule
\multirow{4}{*}{\shortstack[c]{Injection\\defenses}} & StruQ & 88.07 \textpm{} 1.42 & 89.53 \textpm{} 1.84 & 88.77 \textpm{} 0.67 & 98.08 \textpm{} 0.48 & 98.67 \textpm{} 0.18 & 98.38 \textpm{} 0.26 & 97.10 \textpm{} 1.48 & 97.52 \textpm{} 1.42 & 97.31 \textpm{} 1.36 \\
 & SecAlign & 84.16 \textpm{} 3.45 & 87.94 \textpm{} 2.26 & 85.93 \textpm{} 1.11 & 96.24 \textpm{} 0.90 & 98.14 \textpm{} 0.57 & 97.18 \textpm{} 0.63 & 94.91 \textpm{} 3.46 & 97.20 \textpm{} 1.56 & 96.03 \textpm{} 2.47 \\
 & PIGuard & 83.94 \textpm{} 6.60 & 76.90 \textpm{} 11.64 & 79.92 \textpm{} 8.36 & 94.03 \textpm{} 1.81 & 94.20 \textpm{} 2.10 & 94.11 \textpm{} 1.95 & 87.57 \textpm{} 1.19 & 92.43 \textpm{} 2.40 & 89.93 \textpm{} 1.59 \\
 & DRIP & 85.20 \textpm{} 5.31 & 90.02 \textpm{} 3.30 & 87.39 \textpm{} 2.13 & 97.10 \textpm{} 0.77 & 97.19 \textpm{} 0.44 & 97.14 \textpm{} 0.57 & 93.00 \textpm{} 3.08 & 96.04 \textpm{} 1.04 & 94.48 \textpm{} 2.05 \\
\midrule
\multirow{5}{*}{\shortstack[c]{Training\\baselines}} & Flat SFT & 89.66 \textpm{} 2.15 & 87.95 \textpm{} 0.91 & 88.78 \textpm{} 0.86 & 98.26 \textpm{} 0.30 & 98.31 \textpm{} 0.91 & 98.28 \textpm{} 0.61 & 98.25 \textpm{} 0.61 & 98.93 \textpm{} 0.25 & 98.59 \textpm{} 0.23 \\
 & Structured SFT & 89.95 \textpm{} 0.83 & 88.17 \textpm{} 1.72 & 89.04 \textpm{} 0.77 & 98.35 \textpm{} 0.30 & 99.37 \textpm{} 0.10 & 98.86 \textpm{} 0.12 & 98.59 \textpm{} 0.18 & 98.21 \textpm{} 0.85 & 98.40 \textpm{} 0.35 \\
 & R-Drop & 90.72 \textpm{} 0.45 & 86.42 \textpm{} 1.24 & 88.52 \textpm{} 0.80 & 98.35 \textpm{} 0.54 & 99.39 \textpm{} 0.40 & 98.87 \textpm{} 0.42 & 97.51 \textpm{} 0.55 & 99.02 \textpm{} 0.77 & 98.26 \textpm{} 0.31 \\
 & Group DRO & 88.34 \textpm{} 2.50 & 87.57 \textpm{} 2.85 & 87.90 \textpm{} 0.20 & 98.48 \textpm{} 0.46 & 99.05 \textpm{} 0.27 & 98.76 \textpm{} 0.17 & 97.33 \textpm{} 0.88 & 97.14 \textpm{} 2.18 & 97.22 \textpm{} 0.68 \\
 & DPO & 89.06 \textpm{} 1.52 & 89.10 \textpm{} 1.62 & 89.06 \textpm{} 0.12 & 97.66 \textpm{} 0.94 & 98.68 \textpm{} 0.59 & 98.16 \textpm{} 0.25 & 94.58 \textpm{} 3.13 & 96.45 \textpm{} 2.54 & 95.45 \textpm{} 0.77 \\
\midrule
\rowcolor{black!5}
Ours & \textbf{TrustJev} & 90.97 \textpm{} 1.03 & 89.27 \textpm{} 1.14 & 90.11 \textpm{} 0.12 & 99.31 \textpm{} 0.30 & 99.57 \textpm{} 0.25 & 99.44 \textpm{} 0.12 & 99.51 \textpm{} 0.30 & 99.42 \textpm{} 0.00 & 99.46 \textpm{} 0.15 \\
\bottomrule
\end{tabular}%
}
\end{table*}

\subsubsection{Implementation Details}
\label{sec:implementation-details}
TrustJev adapts Qwen3-0.6B with rank-8 LoRA~\cite{hu2022lora} and trainable readouts, using budget ceiling $B(K)=8+\tfrac{1}{2}\log K$ and a 1,536-token sequence limit.
AdamW~\cite{loshchilov2019adamw} uses epochs $(3,2,2)$ and learning rates $(10^{-4},5\times10^{-5},10^{-4})$ for the first, second, and authentication stages, respectively.
Effective batch sizes are two complete source groups for task training and four for authentication.
Adversarial refinement uses 16 candidate attacks per AEB group.
With averaged endpoint cross-entropy weighted at 1, second-stage weights are 1 for stochastic consistency, 0.25 for adversarial cross-entropy, 0.025 each for authority supervision and policy margin, and 0.0075 each for ordinary invariance, preference change, and adversarial invariance.

PIGuard jointly trains its DeBERTa-v3-base encoder and candidate head with MOF replay, averaging scores over overlapping 2,048-token windows without hard blocking.
On IHEval, it receives extended task training and uses the original validation split for checkpoint selection.
We use SecJev-0.8B, Kev-0.8B (Git revision \texttt{54f4f877}), and Jev 1.13.0.
Kev and SecJev use FP32 inference, preserving native LoRA, decision heads, losses, and learning rates during fine-tuning.

Each job uses one 24\,GB RTX 4090.
TrustJev and training baselines select AEB and IHEval checkpoints by validation Balance, evaluating DPO after preference optimization.
On TensorTrust, they use second-epoch checkpoints and calibrate one threshold per method by mean validation Balance.
Kev and SecJev set TensorTrust thresholds using development data.

\subsection{Overall Effectiveness (RQ1)}
\label{sec:overall-effectiveness}

Table~\ref{tab:main-results} reports per-run metrics as means and sample standard deviations for methods evaluated three times.
TrustJev achieves the highest mean Balance, exceeding the strongest baselines by 1.05, 0.57, and 0.87 percentage points on TensorTrust, IHEval, and AEB, respectively.
Mean Utility leads on all datasets and mean Robustness on IHEval and AEB.

TrustJev improves both mean Utility and Robustness over all five training baselines on every dataset, balancing attack resistance with accurate legitimate decisions.
Among injection defenses, DRIP achieves higher Robustness on TensorTrust, but TrustJev has 0.75 points lower Robustness and 5.77 points higher Utility, yielding higher Balance.
On IHEval and AEB, TrustJev exceeds all four defenses in both measures.

Target task fine-tuning makes Kev a competitive decision baseline, tying Flat SFT for the highest baseline Balance on AEB.
TrustJev nevertheless improves both Utility and Robustness over fine-tuned Kev and fine-tuned SecJev on all three datasets.
It also outperforms Jev, the strongest decision system without target task fine-tuning, in both measures throughout.

\subsection{Design Contributions (RQ2)}
\label{sec:design-contributions}

We evaluate individual and grouped component removals from the first training stage, with the same backbone, data, and training budgets.
Isolation removal exposes policy scoring to evidence while retaining semantic positions.
Bound removal uses $r_k=e_k-\bar e$ and removes $2\beta$ from the margin objective.
Paired removal disables auxiliary objectives over related inputs but retains stochastic consistency on the same input.
Authority removal disables policy score cross-entropy, uniformity, and margin supervision.
Both scoring branches, labeled training views, and the remaining objectives are retained.
Each ablated variant is similarly trained with three seeds.

\begin{table}[ht]
\centering
\caption{TrustJev ablations. Balance (\%) shows mean $\pm$ sample SD over three runs. $\times$ denotes removal.}
\label{tab:component-ablations}
\setlength{\tabcolsep}{3pt}
\renewcommand{\arraystretch}{1.05}
\resizebox{\linewidth}{!}{%
\begin{tabular}{cccccccc}
\toprule
\multirow{2}{*}{\textbf{Variant}} & \multicolumn{2}{c}{\textbf{Architecture}} & \multicolumn{2}{c}{\textbf{Supervision}} & \multirow{2}{*}{\textbf{TensorTrust}} & \multirow{2}{*}{\textbf{IHEval}} & \multirow{2}{*}{\textbf{AEB}} \\
\cmidrule(lr){2-3}\cmidrule(lr){4-5}
 & \textbf{Isolation} & \textbf{Bound} & \textbf{Paired} & \textbf{Authority} & & & \\
\midrule
\rowcolor{black!5}
\textbf{Full} &  &  &  &  & 90.11 \textpm{} 0.12 & 99.44 \textpm{} 0.12 & 99.46 \textpm{} 0.15 \\
\midrule
A1 & $\times$ &  &  &  & 87.17 \textpm{} 0.72 & 98.52 \textpm{} 0.21 & 95.18 \textpm{} 0.83 \\
A2 &  & $\times$ &  &  & 88.13 \textpm{} 0.66 & 98.37 \textpm{} 0.11 & 97.24 \textpm{} 0.50 \\
A3 & $\times$ & $\times$ &  &  & 86.42 \textpm{} 0.23 & 98.24 \textpm{} 0.06 & 95.10 \textpm{} 0.47 \\
\midrule
A4 &  &  & $\times$ &  & 88.41 \textpm{} 0.12 & 98.35 \textpm{} 0.16 & 96.22 \textpm{} 0.31 \\
A5 &  &  &  & $\times$ & 86.64 \textpm{} 0.08 & 98.44 \textpm{} 0.13 & 97.21 \textpm{} 0.71 \\
A6 &  &  & $\times$ & $\times$ & 87.05 \textpm{} 1.06 & 98.06 \textpm{} 0.13 & 95.82 \textpm{} 0.31 \\
\midrule
A7 & $\times$ & $\times$ & $\times$ & $\times$ & 85.48 \textpm{} 0.62 & 97.50 \textpm{} 0.23 & 94.40 \textpm{} 0.22 \\
\bottomrule
\end{tabular}%
}
\end{table}

\textbf{Overall ablations.}
Table~\ref{tab:component-ablations} shows that each individual removal lowers mean Balance on all three datasets, supporting contributions from both architectural constraints and dedicated supervision.
Their relative effects vary by task.
Among individual removals, authority supervision removal produces the largest drop on TensorTrust at 3.47 percentage points, while isolation removal has the largest effect on AEB at 4.28 points.
Removing all four components (A7) yields the lowest Balance throughout, indicating that the complete design provides the best overall balance among the evaluated variants.

\begin{figure}[!ht]
    \centering
    \scalebox{1}[0.92]{\includegraphics[width=0.95\linewidth]{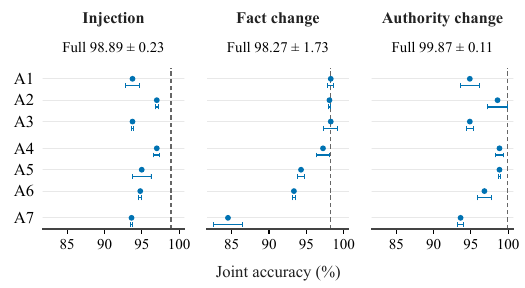}}
    \caption{AEB paired diagnostics for Table~\ref{tab:component-ablations}. Dots and intervals show means and sample standard deviations over three runs. Dashed lines mark Full means, with Full statistics below panel titles. Joint accuracy requires both predictions in a pair to be correct.}
    \label{fig:rq2-aeb-diagnostics}
\end{figure}

\textbf{AEB paired diagnostics.}
Figure~\ref{fig:rq2-aeb-diagnostics} distinguishes component contributions to policy authority and fact responsiveness on the AEB test groups.
Injection pairs clean and attacked inputs with unchanged facts and labels, while fact change pairs inputs with changed facts and labels.
Authority change moves identical override text between evidence and policy, changing the label.
Each diagnostic evaluates all variants on the same input pairs.

Isolation primarily supports injection resistance and authority handling, with its removal lowering their accuracies by 5.14 and 5.01 points while leaving mean fact-change accuracy essentially unchanged.
Removing authority supervision lowers fact-change accuracy by 3.98 points, consistent with leaving room for relevant evidence when policy alone cannot decide.

The architecture's benefit for fact responsiveness depends on supervision.
With both supervision components retained, Full and A3 both average 98.27\% fact-change accuracy.
Without paired and authority supervision, retaining the architecture raises accuracy from 84.50\% (A7) to 93.33\% (A6).

\begin{table}[!ht]
\centering
\caption{Stability by evidence dependence, averaged over three seeds.}
\label{tab:stability-coverage}
\normalfont\normalsize
\setlength{\tabcolsep}{4pt}
\resizebox{\linewidth}{!}{%
\begin{tabular}{llrrrrrrr}
\toprule
\textbf{Task} & \textbf{Type} &
\begin{tabular}[c]{@{}c@{}}\textbf{Utility}\\(\%)\end{tabular} &
\begin{tabular}[c]{@{}c@{}}\textbf{Robustness}\\(\%)\end{tabular} &
\multicolumn{3}{c}{\begin{tabular}[c]{@{}ccc@{}}\multicolumn{3}{c}{\textbf{Percentiles of $s$}}\\\cmidrule(lr){1-3}\textbf{10th} & \textbf{50th} & \textbf{90th}\end{tabular}} &
\begin{tabular}[c]{@{}c@{}}\textbf{Coverage}\\(\%)\end{tabular} &
\begin{tabular}[c]{@{}c@{}}\textbf{Policy}\\\textbf{accuracy}\\(\%)\end{tabular} \\
\midrule
\rowcolor{black!5}
AEB & $P$ & 100.00 & 100.00 & 9.35 & 10.20 & 10.92 & 100.00 & 100.00 \\
AEB & $P+D$ & 99.13 & 99.22 & -14.59 & -14.04 & -13.50 & 0.00 & \textnormal{N/A} \\
\midrule
IHEval Language & $P+D$ & 99.65 & 99.91 & -16.43 & -16.41 & -16.40 & 0.00 & \textnormal{N/A} \\
IHEval Slack & $P+D$ & 98.96 & 99.22 & -16.57 & -16.57 & -16.57 & 0.00 & \textnormal{N/A} \\
\bottomrule
\end{tabular}%
}
\par\smallskip
\begin{minipage}{\linewidth}
\normalfont\footnotesize
$P$ denotes policy-sufficient inputs and $P+D$ evidence-required inputs. Percentiles summarize equally weighted normal and attack views within each seed, pooling AEB tasks. N/A indicates policy accuracy is undefined when coverage is zero.
\end{minipage}
\end{table}

\textbf{Policy stability and evidence dependence.}
Table~\ref{tab:stability-coverage} tests whether learned policy preferences reflect the need for evidence.
With $s=m-2\beta$, coverage is the fraction satisfying $s>0$, which guarantees that evidence corrections cannot change the policy-preferred candidate under Equation~\eqref{eq:authority-margin-bound}.
Policy accuracy measures correctness within this subset.
The analysis covers AEB and IHEval, excluding TensorTrust because its evidence dependence could not be reliably classified.

Policy-sufficient inputs have 100\% coverage and policy accuracy, showing both stable and correct policy preferences.
Evidence-required inputs have zero coverage but retain high Utility and Robustness.
Together with the fact-change diagnostics, these results support policy stability when rules suffice and evidence responsiveness when facts are needed.

\subsection{Robustness Generalization (RQ3)}
\label{sec:robustness-generalization}
\begin{figure}[t]
    \centering
    \scalebox{1}[0.92]{\includegraphics[width=\linewidth]{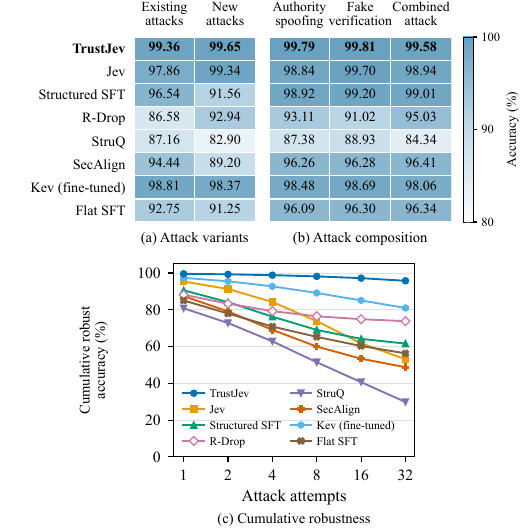}}
    \caption{Robustness under attack variants, composition, and multiple attempts. Panels (a--b) show mean attack accuracy and (c) cumulative robust accuracy for a fixed attack pool on a logarithmic horizontal axis.}
    \label{fig:rq3-robustness}
\end{figure}
RQ3 tests robustness under new attack variants, composition, and multiple attempts.
From RQ1's AEB results, we select the strongest Balance baselines (Flat SFT and fine-tuned Kev), Robustness baseline (R-Drop), and injection defenses by Balance (StruQ and SecAlign), alongside Structured SFT and Jev as references for structured training and commercial decision systems.
Separate AEB sources use the same generator with adjusted task proportions and conditions.
Attacks modify evidence without changing task facts, policies, candidates, or labels.
Panels share sources, and methods receive identical attacks within each panel.

\textbf{Attack variants.}
Figure~\ref{fig:rq3-robustness}(a) compares 16 existing and 16 new attacks per source.
The existing attacks use eight templates, each targeting two incorrect candidates.
New attacks use authority spoofing, fake verification, and candidate lookup through code mappings, individually or combined.
TrustJev maintains high accuracy across libraries, scoring 99.36\% on existing and 99.65\% on new attacks, while Structured SFT, StruQ, and SecAlign have lower accuracy on new attacks.

\textbf{Attack composition.}
Figure~\ref{fig:rq3-robustness}(b) separately reports three conditions within the 16 new attacks per source in (a), with four attacks each for authority spoofing, fake verification, and their combination.
The other four attacks involve candidate lookup and are omitted to focus on the two factors.
All three conditions add 192 tokens to the evidence.
TrustJev scores 99.79\%, 99.81\%, and 99.58\%, respectively.
Under the combined attack, it exceeds Structured SFT, the strongest baseline at 99.01\%, by 0.57 percentage points.

\textbf{Multiple attack attempts.}
Figure~\ref{fig:rq3-robustness}(c) uses a separately constructed fixed pool of 32 attacks per source involving candidate lookup and result spoofing.
For $k\in\{1,2,4,8,16,32\}$, cumulative robust accuracy averages the probability of answering all $k$ attacks correctly when sampled uniformly without replacement.
For $N$ sources, this is $\operatorname{CRA}_m(k)=\frac{100}{N}\sum_{i=1}^{N}\binom{g_{i,m}}{k}/\binom{32}{k}$, where $g_{i,m}$ counts attacks answered correctly by method $m$ on source $i$, and the numerator is zero for $g_{i,m}<k$.
At 32 attempts, TrustJev retains 95.77\% cumulative robust accuracy versus 80.98\% for fine-tuned Kev, the strongest comparator.
Their gap grows from 2.18 percentage points at one attempt to 14.79 at 32.
High single-attempt accuracy does not imply resilience across the attack pool, where TrustJev retains its advantage across multiple attempts.

\subsection{Computational Cost (RQ4)}
\label{sec:computational-cost}
RQ4 compares training and defense approaches sharing TrustJev's backbone under matched hardware and input conditions.
Structured SFT provides the reference cost, while R-Drop, StruQ, SecAlign, and DRIP represent consistency regularization, injection-augmented training, preference optimization, and representation editing, respectively.
We measure three runs with inference batch size 1, reflecting the common workflow setting of making one decision per request.

\begin{table}[t]
    \centering
    \caption{Training cost and peak allocated GPU memory. Training time shows mean $\pm$ sample SD over three runs.}
    \label{tab:rq4-resource-cost}
    \normalfont
    \resizebox{\linewidth}{!}{%
    \begin{tabular}{lrrrrr}
        \toprule
        \multirow{2}{*}{\textbf{Method}} & \multicolumn{2}{c}{\textbf{Training}} & \multicolumn{3}{c}{\textbf{Inference memory (GiB)}} \\
        \cmidrule(lr){2-3}\cmidrule(lr){4-6}
        & \textbf{Time (GPU h)} & \textbf{Peak (GiB)} & \textbf{AEB} & \textbf{IHEval} & \textbf{TensorTrust} \\
        \midrule
        Structured SFT & 0.629 $\pm$ 0.007 & 3.277 & 1.150 & 1.216 & 1.207 \\
        SecAlign & 0.585 $\pm$ 0.003 & 2.331 & 1.150 & 1.216 & 1.208 \\
        DRIP & 0.629 $\pm$ 0.001 & 2.331 & 1.152 & 1.218 & 1.209 \\
        R-Drop & 0.948 $\pm$ 0.003 & 4.021 & 1.150 & 1.216 & 1.207 \\
        StruQ & 0.568 $\pm$ 0.008 & 2.331 & 1.150 & 1.216 & 1.208 \\
        \midrule
        TrustJev & 0.815 $\pm$ 0.015 & 4.022 & 1.150 & 1.216 & 1.207 \\
        \bottomrule
    \end{tabular}%
    }
\end{table}

\begin{figure*}[t]
    \centering
    \includegraphics[width=\textwidth,trim=0 0 0 4bp,clip]{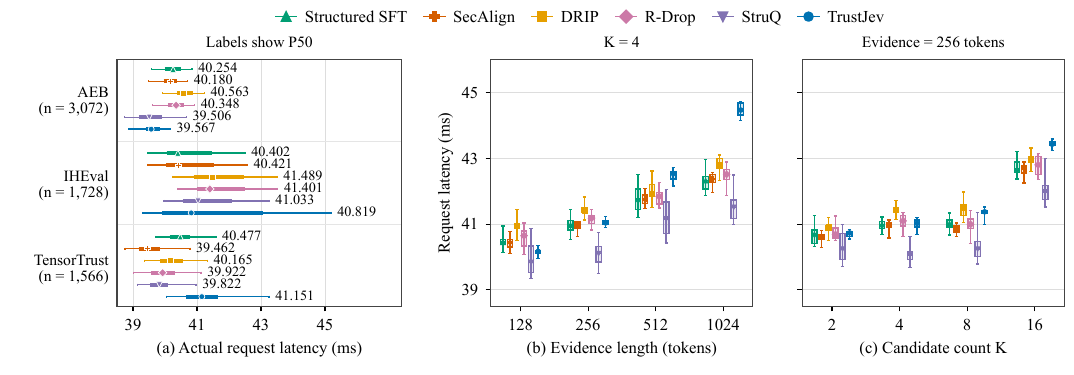}
    \caption{Request latency and input scaling. Panel (a) shows test inputs from three datasets, with P50 labeled. Panel (b) varies evidence length with $K=4$, and panel (c) varies candidate count with 256 evidence tokens. Both sweeps use 8 tokens per candidate and 16 content instances per method and level. Markers indicate P50, thick segments or boxes span P25--P75, and whiskers span P5--P95.}
    \label{fig:rq4-cost-scaling}
\end{figure*}

\textbf{Training and memory.}
Training cost covers initialization, method-specific training and authentication adaptation, reference predictions, validation, and calibration, counting shared stages once per method.
Table~\ref{tab:rq4-resource-cost} shows that TrustJev requires more training time than four baselines but less than R-Drop, while maintaining similar inference memory.
Training takes 0.815 GPU h versus 0.568--0.629 for four baselines and 0.948 for R-Drop.
Peak training memory is 4.022 GiB, close to R-Drop and above the other methods.

\textbf{Request latency.}
We evaluate all test inputs, taking each input's median latency over three executions per seed and then across seeds.
Quantiles summarize these input medians.
Figure~\ref{fig:rq4-cost-scaling}(a) shows median latency near 40--41 ms for TrustJev, comparable to the baselines.
On IHEval and TensorTrust, its P95 is higher at 45.2 and 43.3 ms, versus baseline ranges of 42.5--43.5 and 40.8--41.6 ms, respectively.

\textbf{Input scaling.}
Figure~\ref{fig:rq4-cost-scaling}(b--c) varies evidence length and candidate count separately.
Increasing evidence from 128 to 1,024 tokens raises TrustJev's median latency by 4.3 ms versus 1.7--1.9 ms for the baselines.
Increasing candidate count from 2 to 16 raises its median by 2.7 ms versus 1.7--2.1 ms.
TrustJev has similar median latency on test inputs, with larger increases toward the upper end of the tested evidence lengths.

\subsection{Industrial Alert Triage (RQ5)}
\label{sec:industrial-alert-triage}
To explore industrial performance, we evaluate TrustJev offline on real alerts from nearly ten subsidiaries.
Headquarters triage these alerts for expert intervention, but injected instructions can conceal serious incidents or unnecessarily escalate ordinary alerts.
We construct 178 source cases from 10,349 logs.
The 1,242 evaluation views comprise 581 author-crafted attack views, original alerts, neutral text, and fact-change controls.
Table~\ref{tab:rq5-industrial} reports attack-view classification accuracy, downgrade errors among serious alerts, upgrade errors among ordinary alerts, and latency.
TrustJev achieves 97.65\% accuracy, 5.09 percentage points above direct Jev, with respective error rates of 2.35\% and 2.36\% and P50 latency of 42.11 ms, about one fifth of Jev's.
Rule extraction followed by rules or a text embedding classifier has lower accuracy and higher rates of both errors.
Injected text can cause incorrect fact extraction, propagating errors to either classifier.
TrustJev thus classifies alerts more accurately than the evaluated industrial pipelines and faster than direct Jev.

\begin{table}[t]
\centering
\caption{Industrial alert triage. Accuracy and error rates are percentages. Bold indicates the best result in each column.}
\label{tab:rq5-industrial}
\normalfont\normalsize
\resizebox{\linewidth}{!}{%
\begin{tabular}{lrrrr}
\toprule
\textbf{Method} & \textbf{Accuracy} $\uparrow$ & \textbf{Downgrade error} $\downarrow$ & \textbf{Upgrade error} $\downarrow$ & \textbf{P50 (ms)} $\downarrow$ \\
\midrule
\textbf{TrustJev} & \textbf{97.65} & \textbf{2.35} & 2.36 & 42.11 \\
\midrule
Direct Jev & 92.56 & 16.53 & \textbf{0.00} & 201.47 \\
Extraction + rules & 61.63 & 6.09 & 64.79 & \textbf{0.0075} \\
Extraction + embeddings & 58.04 & 14.07 & 64.79 & 2.80 \\
\bottomrule
\end{tabular}%
}
\end{table}

\section{Lessons Learned}
\label{sec:lessons-learned}

\textbf{Separate policy adherence from fact responsiveness.}
Acceptance requires both injection resistance and correct responses to changed facts, since unchanged answers can reflect policy adherence or evidence neglect.
Our early experiments showed strong authority recognition but weaker fact responsiveness.
Following ML testing practice~\cite{zhang2022mltesting,li2022industrytesting}, our diagnostics pair inputs requiring stable or changed decisions under instruction insertions, fact edits, and authority changes, requiring both endpoints to be correct.

\textbf{Vary fact bindings to discourage answer shortcuts.}
Varying fact positions relative to candidate identifiers discourages positional shortcuts.
In IHEval's Slack task, permuting usernames across records preserves the username set and record identifiers while making the answer follow the shortest username.
Combining these permutations with clean and conflicting instructions supervises both fact responsiveness and resistance to policy overrides.

\textbf{Use executable rules to ground natural-language diagnostics.}
AEB retains natural-language policies and evidence while executable rules label inputs before and after controlled modifications, supporting verifiable attribution of errors to fact use or instruction authority.
Practitioners can similarly label diagnostic cases from rule-governed workflow portions.
These complement realistic evaluations with explicit, independently verifiable expectations for each~modification.

\section{Threats to Validity}
\label{sec:threats-to-validity}

First, implementation, training, and calibration differences between adapted defenses and native decision systems can confound comparisons.
Grouping related views and checking duplicates across splits cannot rule out semantic overlap between cases that inflates generalization estimates.

Second, repeated runs and sample standard deviations leave small performance differences uncertain.
Attacks sharing a source are not independent cases, and fixed attacks do not establish robustness to response-adaptive attacks~\cite{athalye2018obfuscated,tramer2020adaptive}.

Third, evaluation covers one model scale and two or three candidates, assuming application-controlled policies and candidates with trusted instructions separated from external evidence.
AEB underrepresents ambiguous policies and incomplete evidence, and Balance omits unequal operational error costs.
Local latency excludes network and queueing delays, while concurrent serving and full workflow costs require deployment~evaluation.

\section{Related Work}
\label{sec:related-work}

\textbf{Decision interfaces and system defenses.}
Jev, Kev, and SecJev provide typed decisions and candidate scoring for natural-language workflows~\cite{almeida2026jev,palmer2026kev,chen2026secjev}, which external-data injections can redirect~\cite{greshake2023indirect}, including Jev's typed interface~\cite{wu2026decisionhijacking}.
PIGuard detects injections~\cite{li2025piguard}, while CaMeL isolates privileged program generation from untrusted data processing and enforces execution capabilities~\cite{debenedetti2025camel}.
TrustJev complements these controls with bounded evidence influence in candidate~scoring.

\textbf{Instruction and data separation.}
Instruction hierarchy training prioritizes privileged instructions~\cite{wallace2024hierarchy}, Spotlighting marks untrusted content~\cite{hines2024spotlighting}, StruQ pairs protected prompt and data channels with defensive training~\cite{chen2025struq}, and SecAlign favors legitimate over injected responses~\cite{chen2025secalign}.
Instructional Segment Embedding adds instruction-type embeddings~\cite{wu2025ise}, ASIDE rotates data token embeddings~\cite{zverev2026aside}, and DRIP edits data representations and reinforces instructions through residual fusion~\cite{liu2025drip}.
TrustJev scores policies without evidence and bounds corrections by a policy-derived budget, giving a sufficient margin for policy stability as evidence varies.

\textbf{Behavioral supervision and evaluation.}
Metamorphic relations~\cite{chen2018metamorphic} address oracle limitations~\cite{barr2015oracle}, exposing inconsistencies in driving~\cite{tian2018deeptest,zhang2018deeproad} and translation~\cite{he2020sit,sun2020transrepair,he2021purity}.
CheckList tests functionality, invariance, and directionality~\cite{ribeiro2020checklist}, while counterfactual edits change labels to reduce spurious associations~\cite{kaushik2020counterfactual}.
TrustJev couples label-guided stability and change with authority supervision, distinguishing policy-sufficient and evidence-required decisions.
BIPIA tests external content injections~\cite{yi2025bipia}, InjecAgent tool response injections~\cite{zhan2024injecagent}, and AgentDojo agent task completion and security during tool use~\cite{debenedetti2024agentdojo}.
AEB separates injection resistance from evidence neglect in finite candidate decisions using executable labels and controlled fact and authority edits.

\section{Conclusion}
\label{sec:conclusion}

TrustJev combines authority isolation, bounded evidence corrections, and behavioral supervision for policy-guided candidate decisions.
Its 0.6B implementation leads 14 baselines in mean Balance on three datasets and withstands all 32 fixed-pool AEB attacks on 95.77\% of sources versus the strongest comparator's 80.98\%.
These gains increase training cost and latency growth with evidence length, while typical inference latency stays comparable and memory nearly unchanged.

\bibliographystyle{IEEEtran}
\bibliography{ref}

\end{document}